\documentclass{ecai}
\usepackage{times}
\usepackage{graphicx}
\usepackage{latexsym}

\usepackage{booktabs}

\usepackage{mwe}
\usepackage{tabularx}
\usepackage{enumitem}
\setlist[itemize]{leftmargin=*}
\usepackage{soul}
\usepackage{algorithm}
\usepackage{stackengine}
\usepackage{algpseudocode}
\usepackage{siunitx}
\usepackage{bm}
\usepackage{balance}
\usepackage{textcomp}
\usepackage{listings}
\usepackage{multirow}
\usepackage{pdfpages}
\usepackage{balance}
\usepackage{setspace}
\usepackage[framemethod=tikz]{mdframed}
\usepackage[justification=centering]{caption}
\usepackage{subfigure}
\graphicspath{ {images/} }
\usepackage{afterpage}
\usepackage{float}
\usepackage{url}
\usepackage{longtable}
\usepackage{dirtytalk}
\usepackage{amsmath}
\usepackage[export]{adjustbox}
\usepackage{amssymb}
\usepackage{mathtools}
\DeclarePairedDelimiter\abs{\lvert}{\rvert}%

\usepackage{array}
\usepackage{tabu}
\usepackage{tikz}
\usetikzlibrary{calc}
\usetikzlibrary{matrix}
\usetikzlibrary{shapes}
\usetikzlibrary{positioning}
\usetikzlibrary{shapes,arrows}
\tikzstyle{decision} = [diamond, draw, fill=blue!20, 
    text width=4.5em, text badly centered, node distance=3cm, inner sep=0pt]
\tikzstyle{block} = [rectangle, draw, fill=blue!20, 
    text width=5em, text centered, rounded corners, minimum height=4em]
\tikzstyle{line} = [draw, -latex']
\tikzstyle{cloud} = [draw, ellipse,fill=red!20, node distance=3cm,
    minimum height=2em]

\definecolor{black}{rgb}		{0.0, 0.0, 0.0}
\definecolor{white}{rgb}		{1.0, 1.0, 1.0}
\definecolor{yellow}{rgb}		{1.0, 1.0, 0.8}
\definecolor{red}{rgb}			{0.6, 0.0, 0.2}
\definecolor{blue}{rgb}		{0.0, 0.2, 0.5}
\definecolor{green}{rgb}		{0.6, 0.8, 0.8}
\definecolor{dark_green}{RGB} {0, 140, 0}
\definecolor{gold}{rgb}		{0.6, 0.4, 0.1}
\definecolor{grey}{RGB}{0,0,0}
\definecolor{purple}{RGB}{128,0,128}
\definecolor{sl_blue}{RGB}{47, 60, 105}
\definecolor{orange}{RGB}{255,165,0}
\definecolor{Gray}{gray}{0.85}

\usepackage{marvosym}

\makeatletter
\newif\if@anonymize

\@anonymizefalse  

\if@anonymize
  \newcommand{\highlight@DoHighlight}{
    \fill [outer sep = -15pt, inner sep = 0pt, color=black]
          ($(begin highlight)+(0,8pt)$) rectangle ($(end highlight)+(0,-3pt)$) ;
  }

  \newcommand{\highlight@BeginHighlight}{
    \coordinate (begin highlight) at (0,0) ;
  }

  \newcommand{\highlight@EndHighlight}{
    \coordinate (end highlight) at (0,0) ;
  }

  \newdimen\highlight@previous
  \newdimen\highlight@current
  \newlength{\item@width}

  \DeclareRobustCommand*\anonymize{%
    \SOUL@setup
    \def\SOUL@preamble{%
      \begin{tikzpicture}[overlay, remember picture]
        \highlight@BeginHighlight
        \highlight@EndHighlight
      \end{tikzpicture}%
    }%
    \def\SOUL@postamble{%
      \begin{tikzpicture}[overlay, remember picture]
        \highlight@EndHighlight
        \highlight@DoHighlight
      \end{tikzpicture}%
    }%
    \def\SOUL@everyhyphen{%
      \discretionary{%
        \SOUL@setkern\SOUL@hyphkern
        \SOUL@sethyphenchar
        \tikz[overlay, remember picture] \highlight@EndHighlight ;%
      }{%
      }{%
        \SOUL@setkern\SOUL@charkern
      }%
    }%
    \def\SOUL@everyexhyphen##1{%
      \SOUL@setkern\SOUL@hyphkern
      \settowidth{\item@width}{##1}%
      \makebox[\item@width]{}%
      \discretionary{%
        \tikz[overlay, remember picture] \highlight@EndHighlight ;%
      }{%
      }{%
        \SOUL@setkern\SOUL@charkern
      }%
    }%
    \def\SOUL@everysyllable{%
      \begin{tikzpicture}[overlay, remember picture]
        \path let \p0 = (begin highlight), \p1 = (0,0) in \pgfextra
          \global\highlight@previous=\y0
          \global\highlight@current =\y1
        \endpgfextra (0,0) ;
        \ifdim\highlight@current < \highlight@previous
          \highlight@DoHighlight
          \highlight@BeginHighlight
        \fi
      \end{tikzpicture}%
      \settowidth{\item@width}{\the\SOUL@syllable}%
      \makebox[\item@width]{}%
      \tikz[overlay, remember picture] \highlight@EndHighlight ;%
    }%
    \SOUL@
  }
\else
  \newcommand{\anonymize}[1]{#1}
\fi

\begin{document}

\title{The Automated Inspection of Opaque Liquid Vaccines}

\author{G. Palmer$^1$ \and B. Schnieders\textsuperscript{\Cross,} $^1$, \and R. Savani$^1$ \and K. Tuyls\institute{University of Liverpool, UK, contact author: G.J.Palmer@liverpool.ac.uk} \and J. Fossel \and H. Flore\institute{HAL Allergy Group, Netherlands.}}


\maketitle
\bibliographystyle{ecai}

\begin{abstract}
In the pharmaceutical industry the screening of opaque vaccines containing suspensions is currently a manual task carried out by trained human visual inspectors. We show that deep learning can be used to effectively automate this process. A moving contrast is required to distinguish anomalies from other particles, reflections and dust resting on a vial's surface. We train 3D-ConvNets to predict the likelihood of 20-frame video samples containing anomalies. Our unaugmented dataset consists of hand-labelled samples, recorded using vials provided by \anonymize{the HAL Allergy Group}, a pharmaceutical company. We trained ten randomly initialized 3D-ConvNets to provide a benchmark, observing mean AUROC scores of 0.94 and 0.93 for positive samples (containing anomalies) and negative (anomaly-free) samples, respectively. Using Frame-Completion Generative Adversarial Networks we: (i) introduce an algorithm for computing saliency maps, which we use to verify that the 3D-ConvNets are indeed identifying anomalies; (ii) propose a novel self-training approach using the saliency maps to determine if multiple networks agree on the location of anomalies. Our self-training approach allows us to augment our data set by labelling 217,888 additional samples. 3D-ConvNets trained with our augmented dataset improve on the results we get when we train only on the unaugmented dataset.
\end{abstract}

\section{Introduction}

One of the challenges faced within the pharmaceutical industry is the screening of liquid vaccines (also referred to as suspensions). A visual inspection process is required to ensure that opaque suspensions are free of undesirable particles, since aggregates are believed to cause unwanted immunogenic responses~\cite{das2012protein,nygaard2004capacity,Gretchen2016}. Screening suspensions is currently a manual task carried out by trained human visual inspectors. The inspection process requires the content of each vial to be shaken up in order to identify anomalies, which are frequently only visible for an instant\footnote{The challenging nature of the task can be seen in the following video: \url{https://youtu.be/S1IapmRl9H0}}. Human inspectors must remain focused while processing large batches of vials. Therefore, while manual inspection is effective, an automated approach offers significant potential towards a reliable cost-effective inspection. An automated solution could prove invaluable during an epidemic, allowing pharmaceutical companies to increase the production rate of vaccines containing suspensions without having to make a compromise regarding product integrity. In contrast, recruiting and training new visual inspectors would delay the role-out of much needed~medicines. 

In recent years there have been significant advances within the field of automated image and video classification, with \emph{deep learning} techniques utilizing \emph{Convolutional Neural Networks} (ConvNets) setting new standards. In this paper we show that deep learning can be used to detect non-desirable particles within vaccines consisting of an \emph{opaque} liquid. Our contributions can be summarized as follows:

\noindent{\textbf{1)}} We outline our process for constructing a video dataset using vaccines supplied by \anonymize{the HAL Allergy Group}. We built an automated vial rotator (AVR) for inducing the moving contrast necessary to identify anomalies~\cite{hprinz}. However, recordings suffer from \emph{motion blur} over the first 20 -- 40 frames due to particles moving at a high velocity after the vial is spun (See Fig. \ref{fig:motion_blur}). Due to motion blur \emph{good} particles within the vials can appear elongated, making them hard to distinguish from anomalies. To evaluate the extent to which motion blur affects classification accuracy we split recorded samples into segments consisting of 20 frames. We hand-labelled the segments based on the presence of anomalies, enabling us to construct a dataset of 14k training and 6k evaluation samples, derived from 160 vials. 

\noindent{\textbf{2)}} We empirically evaluate the ability of 3D-ConvNets~\cite{ji20133d} to detect anomalies using our initial dataset, observing average AUROC (Area Under the ROC Curve) scores of 0.94 and 0.93 for positive samples (containing anomalies) and negative (anomaly-free) samples, respectively. We also find evidence that excluding samples with motion blur improves classification accuracy. 

\noindent{\textbf{3)}} We introduce an algorithm for computing saliency maps to verify that predictions are based on the presence of anomalies. For this we use Frame-Completion Generative Adversarial Networks (FC-GANs)~\cite{iizuka2017globally} to identify frame regions that impact predictions. We conduct a qualitative evaluation of the saliency maps, finding predictions are predominately relying on the correct input features.

\noindent{\textbf{4)}} Due to the small number of training samples the 3D-ConvNets over-fit after 100 epochs. To address this issue we use self-training (bootstrapping) for augmenting our dataset~\cite{yarowsky1995unsupervised}, incorporating the FC-GANs-based saliency maps into a multi-classifier voting system to automatically label additional training samples. Upon optimizing 3D-ConvNets using the additional samples we observe improved AUROC scores of 0.96 for positive and negative evaluation samples. 

The remainder of the paper proceeds as follows: first we discuss the related work (Section~\ref{sec:related_work}), followed by the visual particle inspection challenge and the relevant background literature (Section~\ref{sec:background}). We subsequently outline our dataset construction process (Section~\ref{sec:dataset}) and benchmark the ability of 3D-ConvNets to detect anomalies in suspensions (Section~\ref{sec:3D-ConvNets}). In Section~\ref{sec:verifcation} we introduce FC-GANs as a means to produce saliency maps, which serve a double purpose: (i)~we use the saliency maps to verify that predictions are based on the presence of anomalies; (ii)~we propose a novel self-training technique that incorporates the saliency maps into a multi-classifier voting system. We find that networks trained using the augmented self-training dataset outperform 3D-ConvNets trained with the unaugmented dataset (Section~\ref{sec:Self-training}). We consider future work in Section~\ref{sec:FutureWork}, and conclude the paper in Section~\ref{sec:conclusion}.

\section{Related Work} \label{sec:related_work}

Past efforts towards automating particle inspection have relied on segmentation methods for tracking and classifying potential anomalies individually using an adaptive sampling strategy~\cite{zhang2018automated}. However, for the evaluated product capturing eight sequential images for each vial is sufficient for classification. In contrast, the anomalies within the product discussed in this paper are often obscured due to high opacity. Therefore, longer image sequences are required to detect anomalies. Furthermore, the large number of \emph{good} particles increases the computational burden for segmenting and identifying each potential defect. Recently Tsay and Li~\cite{tsay2019automating} showed that deep learning can be used to detect faults in lyophilized (immobile) drug products. Faults were detected with 85-90\% accuracy using samples consisting of six images at 90 degree rotations. The authors encountered challenges with the network overfitting due to limited amounts of training data. To mitigate overfitting transfer learning was used. In contrast, we turn to self-training in Section~\ref{sec:Self-training}. Zhao et al.~\cite{zhao2018joint} investigate a mobile product using a single-frame Faster-RCNN network combined with clustering to determine the target motion area. However, the authors find that single image classification is unreliable for this task. Therefore, due to our product requiring a moving contrast for identifying anomalies we turn to 3D Convolution Neural Networks (3D-ConvNets) for our evaluation, and propose FC-GANs-based saliency maps in Section~\ref{sec:verifcation} for determining the target motion area.

\section{Background} \label{sec:background}

In this section we first summarize the challenges of visually inspecting opaque liquid vaccines containing suspensions, before discussing the techniques drawn upon to overcome them. 

\subsection{Visual Particle Inspection Challenges} \label{sec:visual_inspection_challenge}

\begin{figure}[h]
\centering
\subfigure[Reflection]{\includegraphics[width=0.45\columnwidth]{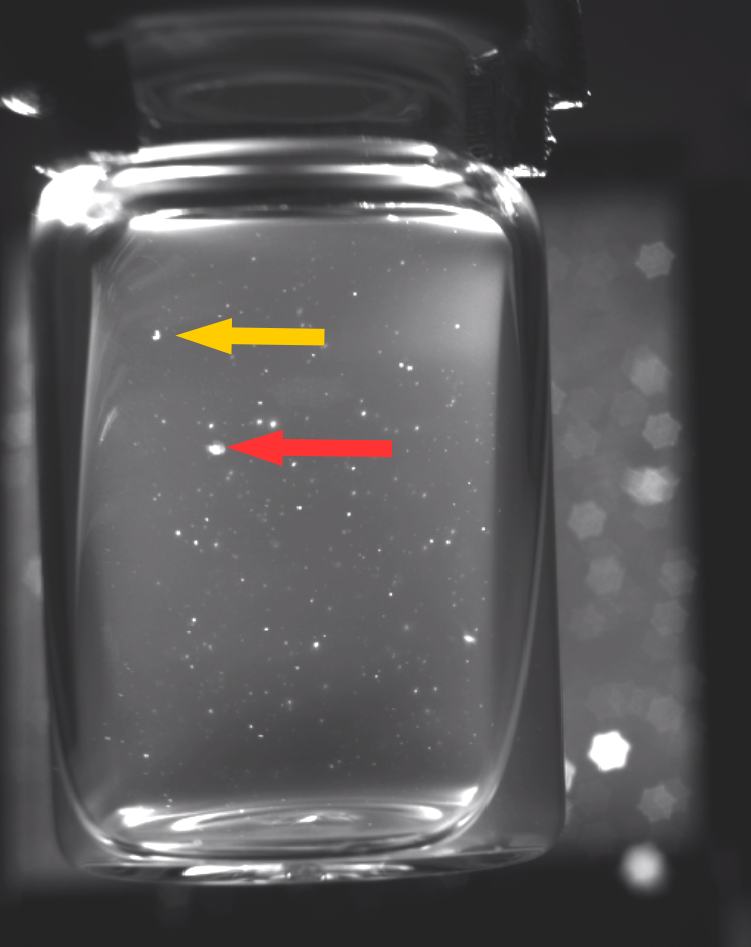}} \hspace{3mm}
\subfigure[Bubbles]{\includegraphics[width=0.45\columnwidth]{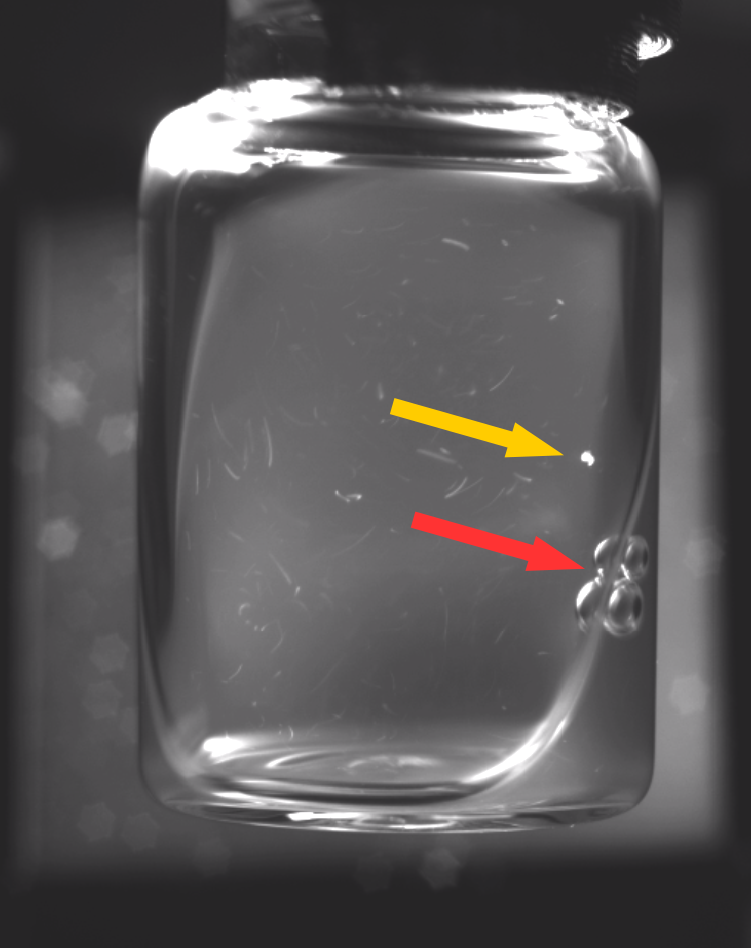}}
\subfigure[Dust Particles]{\includegraphics[width=0.45\columnwidth]{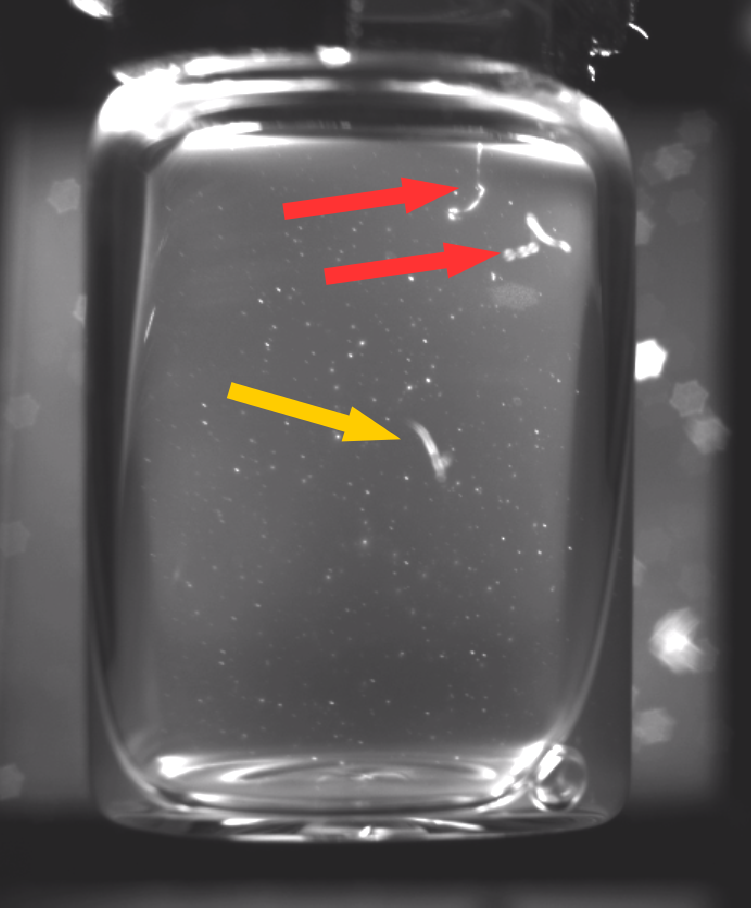}} \hspace{3mm}
\subfigure[Dust Particle]{\includegraphics[width=0.45\columnwidth]{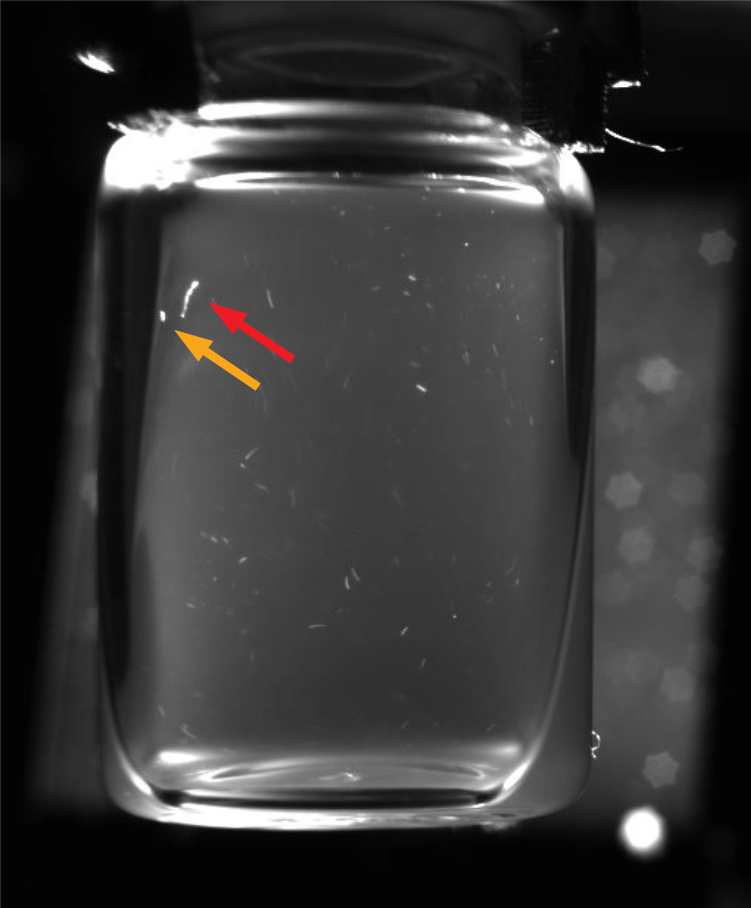}}
\caption{Yellow arrows point to anomalies within the liquid vaccines, red arrows point to the labelled entity.}
\label{fig:vials}
\end{figure}

Correct lighting conditions are a prerequisite for identifying anomalies within liquid vaccines, since, due to a lack of contrast, their identification under natural lighting conditions is currently infeasible for human or automated visual inspection. A light intensity must be found that is sufficient for illuminating the vial while providing a moving contrast to identify the smallest particles~\cite{hprinz}. To obtain a moving contrast human inspectors shake the vials to induce a swirl during manual inspections. Automated approaches meanwhile, such as the semi-automatic Seidenader V90+ inspection machine, use servo motors to stir up particles inside the vials. However, this approach has a side-effect of creating bubbles within the liquid. Furthermore, light reflections and dust particles resting on the outer surface of the vials can often be mistaken for anomalies within the liquid~\cite{hprinz}. Opaque suspensions further increase the task difficulty by obscuring anomalies, which as a result are frequently only briefly visible. Figure \ref{fig:vials} depicts some of these challenges. 

\subsection{Convolutional Neural Networks}
Convolutional neural networks (ConvNets) represent the current state of the art for image classification tasks~\cite{huang2017densely,he2016deep}. Their strength lies in their large learning capacity, which can be adjusted through changing the network's depth and breadth~\cite{krizhevsky2012imagenet}. ConvNets take advantage of assumptions regarding the location of pixel dependencies within images, reducing the number of weighted connections compared to a fully-connected neural network~\cite{krizhevsky2012imagenet}. Traditional ConvNet architectures consist of multiple linear convolution and pooling layers stacked up on top of each other followed by fully connected layers preceding the classification layer~\cite{sun2016fish}. The convolutional layers are banks of filters which are convoluted with an input to produce an output map~\cite{iizuka2017globally}. A non-linear activation function is then applied to the output map such as the Rectified Linear Unit (ReLU)~\cite{nair2010rectified}. 

\subsection{Video Classification}
The moving contrast required for detecting anomalies means networks must be able to process a temporal dimension. Two methods for coping with this additional dimension are \emph{Long Short-Term Memory} (LSTM) cells for an arbitrary length history~\cite{hochreiter1997long} and \emph{3D-ConvNets} using three dimensional convolutional layers~\cite{ji20133d}. The filter size within each convolutional layer is therefore set to a defined height $H$, width $W$, color channel size $C$ and length $T$, representing the filter's length along the temporal dimension~\cite{ji20133d,karpathy2014large}:  $H \times W \times C \times T$. In this paper, we use 3D-ConvNets.

\subsection{Greying the Black-Box} \label{sec:background:gregying}

Despite being a black-box based technique, deep learning models are increasingly deployed in safety-critical systems~\cite{pei2017deepxplore}. While mis-classification of edge cases cannot be ruled out, there have been efforts to ``grey out the black-box''. DeepXplore for instance systematically evaluates deep learning architectures, using a neuron coverage metric to measure the number of rules that are exercised by a set of network inputs, thereby identifying erroneous behaviours~\cite{pei2017deepxplore}. Alternatively, saliency maps can be computed to identify salient features within network inputs, using either gradient or perturbation-based saliency methods~\cite{greydanus2017visualizing}. In Section \ref{sec:verifcation} we use saliency maps to verify that 3D-ConvNets are identifying anomalies within the vials. However, unlike Greydanus et al.~\cite{greydanus2017visualizing} we replace sub-regions in the input frames with a realistic anomaly free content to find the regions with the biggest impact on the prediction. We create the replacement anomaly free content with Generative Adversarial Networks~(GANs)~\cite{goodfellow2014generative}.

\subsection{Generative Adversarial Networks}

Goodfellow et al. \cite{goodfellow2014generative} proposed GANs for capturing the distribution of a dataset. GANs consist of two adversarial networks: a generative model $G$ and a discriminator $D$. The networks play a game where $D$ is trained to distinguish dataset samples from those originating from $G$, while $G$ learns to maximize the probability of fooling $D$. The discriminator's loss is used to guide the optimization of $G$. GANs have been used to capture the distribution of a number of dataset types, including images~\cite{mathieu2015deep}, videos~\cite{mathieu2015deep,vondrick2016generating}, text~\cite{fedus2018maskgan}, 3D models~\cite{wu2016learning} and even pharmaceutical drugs~\cite{kadurin2017drugan}. Furthermore GANs deliver impressive results when tasked with completing an image with a masked area. Iizuka et al.~\cite{iizuka2017globally} trained an image completion network tasked with fooling two discriminators: a local discriminator focusing on the output produced for the masked area, and a global discriminator that processed the entire image. The resulting generator is capable of removing objects within an image, replacing the extraction area with realistic content. We want to achieve a similar outcome within our vial samples in order to predict the location of the anomalies within our vials, which we discuss in more detail in Section \ref{sec:verifcation}.

\subsection{Self-Training} \label{sec:background:self-training}

Self-training was introduced by David Yarowsky~\cite{yarowsky1995unsupervised} as a method for word-sense disambiguation, where an initial classifier is trained using only a small set of labelled samples. The learned rules are used to assign labels to unlabelled samples, allowing a fresh classifier to be trained using a larger dataset. This bootstrapping approach is useful for tasks where gathering large amounts of labelled data is infeasible due to the cost associated with hand-labelling samples~\cite{bank2018improved}. However, automatic labelling requires considerations regarding reducing the impact of \emph{noisy} labels resulting from mis-classification~\cite{xiao2015learning}. A multi-classifier voting system with a defined level of strictness can reduce the number of noisy labels~\cite{rajendran2018something}. For our current task we also have an insufficient number of labelled samples, despite investing a considerable amount of time into the dataset construction process described below. In Section \ref{sec:Self-training} we use self-training to address this issue.

\section{Dataset Construction} \label{sec:dataset}

In this section we describe how we recorded and labelled our dataset.

\begin{figure}[h]
\centering
\subfigure[]{\includegraphics[width=0.48\columnwidth]{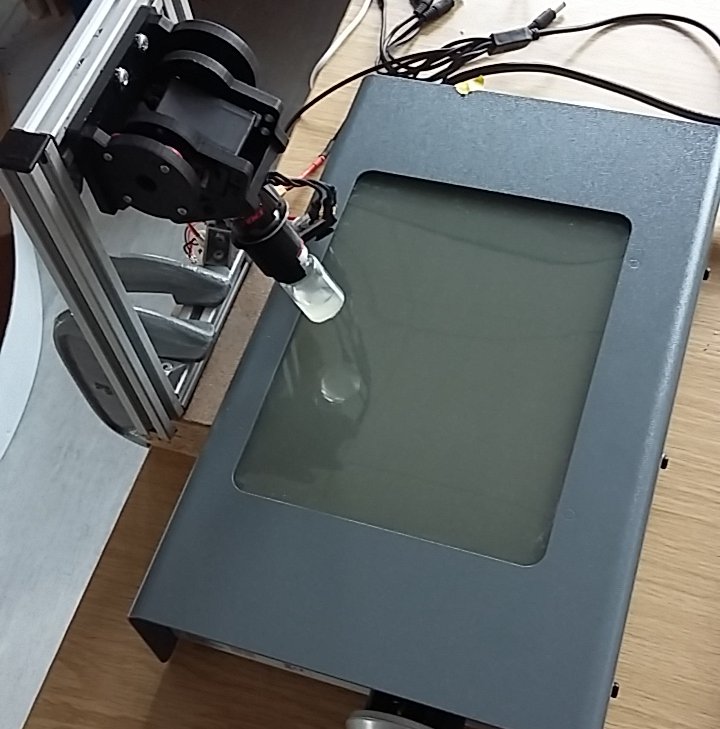}} \hspace{1mm}
\subfigure[]{\includegraphics[width=0.48\columnwidth]{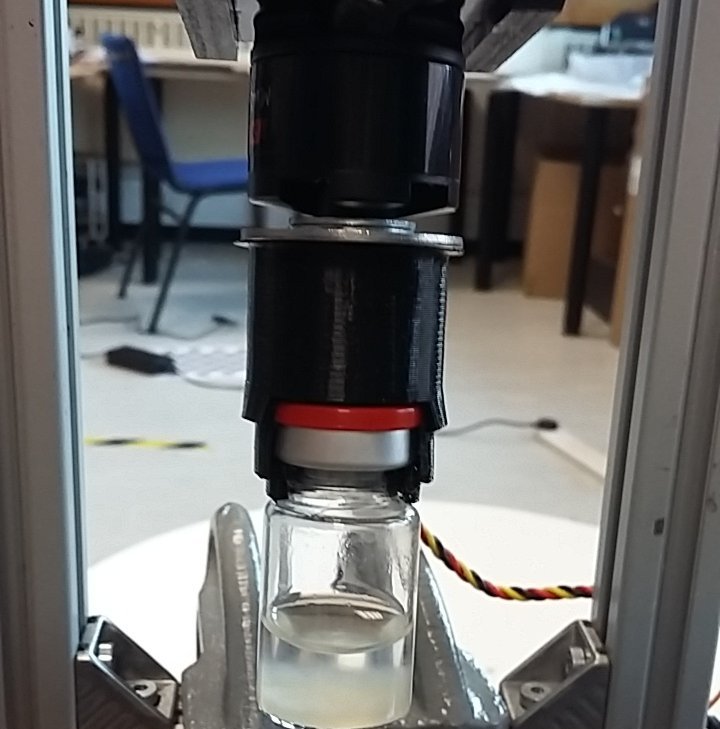}}
\caption{Photos of the \textit{Automated Vial Rotator}}
\label{fig:photos_avr}
\end{figure}

\noindent{\textbf{Equipment:}} \anonymize{The HAL Allergy Group} provided the 160 vials of product type \anonymize{P02U40} that we used for recording our dataset, and financed an AlliedVision MANTA G-235B POE monochrome network camera and a CCS TH2-51/51-SW Compact homogeneous LED back-light. We implemented an Arduino controlled Automated Vial Rotator (AVR) to ensure that the recordings are standardised. Using a Brushless Motor Emax MT2213 935Kv our AVR is capable of inducing a swirl inside a vial to stir up the contents. Inspections of upright standing containers have been shown to have poor detection rates~\cite{hprinz}. We therefore added a servo for adjusting the inspection angle to increase the recorded surface area. This addition allows us to take full advantage of the LED back-light to narrow the camera's aperture sufficiently and increase the depth of focus. While our current AVR is not intended for a pharmaceutical production workflow (for which efficient conveyor belt solutions already exists), it does provided a means through which to record a dataset in the setting of our research institution. We provide photos of our AVR in Figure \ref{fig:photos_avr}.

\begin{figure}[h]
\centering
\subfigure[]{\includegraphics[width=0.45\columnwidth]{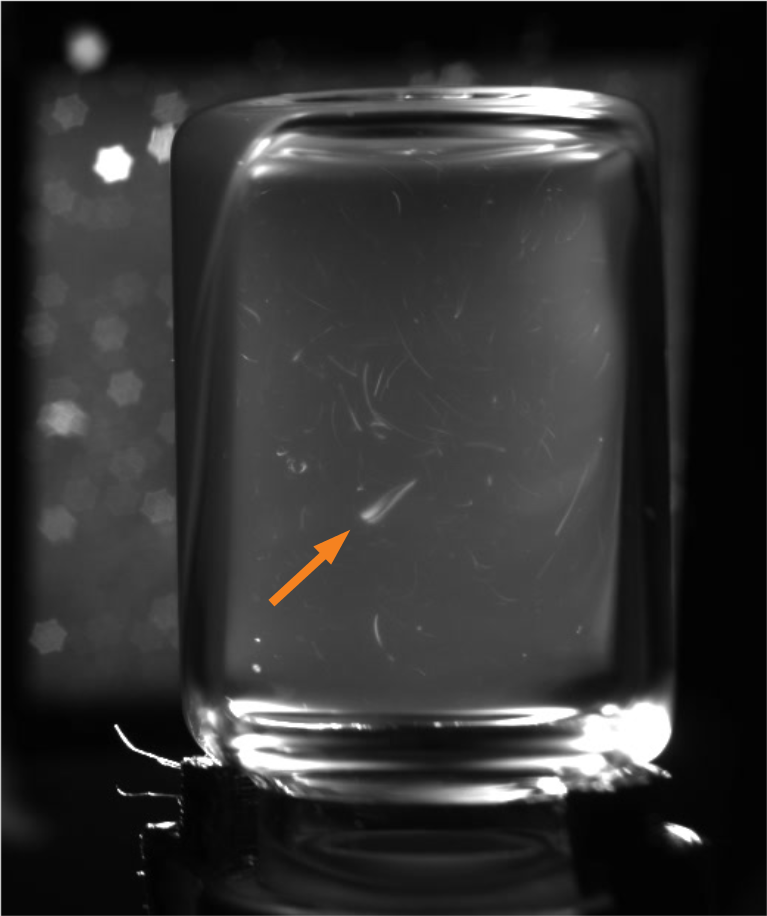}} \hspace{3mm}
\subfigure[]{\includegraphics[width=0.45\columnwidth]{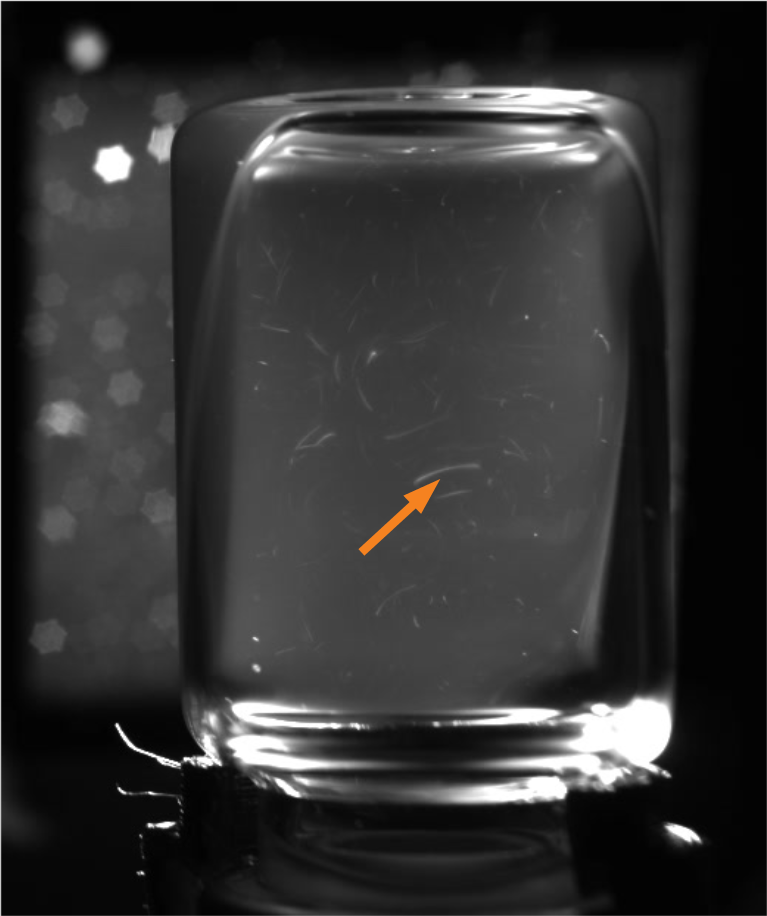}}
\caption{Two examples of frames suffering from motion blur. Arrows point to anomalies within the vials.}
\label{fig:motion_blur}
\end{figure}

\noindent{\textbf{Motion Blur:}} One of the challenges regarding tuning the camera prior to recording the vials, was to find a depth of field that provides a sharp focus for all the particles within the suspensions. This means that the aperture size has to be narrowed to enable a sufficiently deep depth of field. A smaller aperture requires longer shutter speeds in order for sufficient light to reach the camera’s sensor. Despite increasing the light emitted by our strobe to the maximum setting, we are only able to record using 25 fps, and as a result the initial 20 -- 40 frames from each recording suffer from motion blur (Examples are provided in Figure \ref{fig:motion_blur}). Therefore, due to particles' increased velocity after the vial is rotated using the motor, even \emph{good} particles appear elongated during the initial frames of each recording.

\noindent{\textbf{Recording process:}} Prior to recording our dataset the vials were split into three categories based on the difficulty in manually locating anomalies: 66 \emph{Anomaly Free} (AF), 43 \emph{Easy Rejects} (ER) and 51 \emph{Challenging Rejects} (CR). From each category 20 vials were set aside for recording an evaluation set. We recorded 2k training and 1k evaluation videos per category. We were uncertain of the impact repeated exposure to the AVR's forces would have on the integrity of each vial's content. Therefore, each vial was visually inspected prior to being recorded, to ensure it still belonged to the designated category. Upon completing the recording process we applied a pre-processing script to our recordings, using background subtraction to establish the active region with regards to floating particles within each frame. This allowed us to discard static particle free areas of each recording. We subsequently down-sampled and cropped each sample's $160$ frames to a $100 \times 100$ pixel region based on the upper left most active pixel coordinate.

\noindent{\textbf{Labelling:}} We conducted an initial trial run upon completing the steps outlined above, with limited success. We believe there are two reasons why the 3D-ConvNets struggle to learn to detect anomalies directly from the 160-frame sequences:

\begin{enumerate}
    \item despite our efforts dust particles frequently attached themselves to the vials, representing a potential confounding factor given the limited sample size;
    \item we hypothesize that the 3D-ConvNets are sensitive towards the velocity of the particles in the liquid, i.e., excluding samples suffering from motion blur will improve the classification accuracy. 
\end{enumerate}

To test this hypothesis we hand-labelled 14,000 training samples ($\frac{1}{2}$~AF, the other $\frac{1}{2}$~ER \& CR) consisting of 20-frame sequences, plus an additional 6000 samples from our evaluation recordings ($\frac{1}{3}$~ER, $\frac{1}{3}$~CR, $\frac{1}{3}$~AF). Extracting 20-frame sequences from 160-frame recordings allows frames belonging to the same vial to be distributed across both \emph{positive}  (containing anomalies) and \emph{negative} (anomaly-free) labels when the anomaly is only visible within a subset of frames. We assign one video level label -- \emph{positive} or \emph{negative} -- to each 20-frame sequence extracted from the 160-frame recording. Each 20-frame sequence is treated as an independent sample, where an anomaly appearing in 1 out 20 frames is a sufficient condition to label the 20-frame sequence as positive. This additional step reduces the likelihood of the networks learning to classify based on confounding factors such as dust particles, bubbles within the liquid and reflections. However, we note that in practice the classifications from each 20-frame sample extracted from a recording could be aggregated, with one 20-frame sample receiving a positive classification being sufficient to reject a vial.

\section{Evaluation of 3D-ConvNets} \label{sec:3D-ConvNets}

Upon completing the manual labelling process we train ten randomly initialized 3D-ConvNets on our dataset. Each network receives samples consisting of $100 \times 100 \times 20$ pixel values as inputs. The networks consist of four 3D convolutional layers with 32, 64, 64 and 128 filters, a fully-connected layer with 1024 nodes and finally a Sigmoid output layer. Adam~\cite{kingma2014adam} is used to minimize the cross entropy loss 
$H_{y'}(y) = - \sum_{i=1}^{2} y_i' log(y_i)$, where $y_i$ represents the prediction, $y_i'$ the true data label, and there are two classes, $i=1,2$.

We achieve a mean prediction accuracy of 85\% across the ten trained networks. However, a closer look at the accuracy and loss conditioned on the frame-range during which the sample was extracted reveals interesting insights. We observe that due to motion blur predictions made for frames extracted between time-steps 0 and 20 are generally poor (78.6\%). Meanwhile, for positive samples (ER \& CR) the highest accuracy / lowest loss is observed between time-steps 40 to 100, with 89.6\% accuracy for ER and 82.2\% for CR. For negative (anomaly free) samples we observe an increase in correct classifications and lower losses in frames with less movement. However, 88.7\% is the highest percentage of correct predictions across all evaluation sets, achieved between frames 80 and 100. Therefore, sufficient motion is required to distinguish anomalies from confounding factors. These findings support our hypothesis from Section \ref{sec:dataset}, that classification accuracy is dependent on the velocity of the particles. Therefore, our models are able to more accurately classify samples not suffering from motion blur, where mis-classification can occur as a result of good particles appearing elongated. We illustrate the average loss scores for frame ranges in Table \ref{tbl:accuracy} in Section \ref{sec:Self-training}, where we compare the performance of our initial classifiers with those optimised via self-training.

\begin{table*}[h]
        \centering
        \resizebox{\textwidth}{!}{%
        \bgroup
        \tabulinesep=1.5mm
        \begin{tabu}{p{1.7cm}cccccccccc}
            \toprule
            \textbf{Time-step:} & \textbf{1} & \textbf{2} & \textbf{3} & \textbf{4} & \textbf{5} & \textbf{6} & \textbf{7} & \textbf{8} & \textbf{9} & \textbf{10} \\
            \toprule
            \textbf{Ground Truth:}
            & 
            \raisebox{-.5\height}{\includegraphics[height=1.8cm]{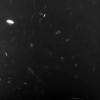}}
            & 
            \raisebox{-.5\height}{\includegraphics[height=1.8cm]{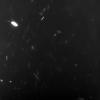}}            
            & 
            \raisebox{-.5\height}{\includegraphics[height=1.8cm]{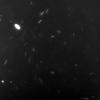}}
            & 
            \raisebox{-.5\height}{\includegraphics[height=1.8cm]{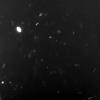}}
            &
            \raisebox{-.5\height}{\includegraphics[height=1.8cm]{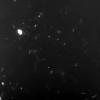}}
            & 
            \raisebox{-.5\height}{\includegraphics[height=1.8cm]{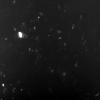}}
            & 
            \raisebox{-.5\height}{\includegraphics[height=1.8cm]{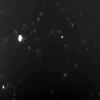}}            
            &
            \raisebox{-.5\height}{\includegraphics[height=1.8cm]{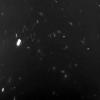}}
            & 
            \raisebox{-.5\height}{\includegraphics[height=1.8cm]{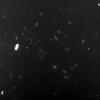}}
            & 
            \raisebox{-.5\height}{\includegraphics[height=1.8cm]{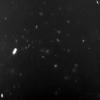}}            
            \\
            \hline
            \textbf{FC-GANs Example Input:}
            & 
            \raisebox{-.5\height}{\includegraphics[height=1.8cm]{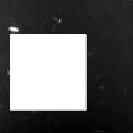}}
            & 
            \raisebox{-.5\height}{\includegraphics[height=1.8cm]{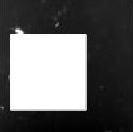}}            
            & 
            \raisebox{-.5\height}{\includegraphics[height=1.8cm]{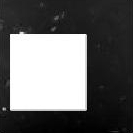}}
            & 
            \raisebox{-.5\height}{\includegraphics[height=1.8cm]{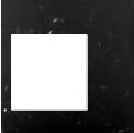}}
            &
            \raisebox{-.5\height}{\includegraphics[height=1.8cm]{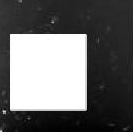}}
            & 
            \raisebox{-.5\height}{\includegraphics[height=1.8cm]{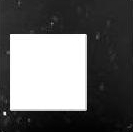}}
            & 
            \raisebox{-.5\height}{\includegraphics[height=1.8cm]{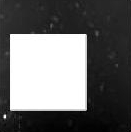}}            
            &
            \raisebox{-.5\height}{\includegraphics[height=1.8cm]{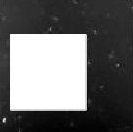}}
            & 
            \raisebox{-.5\height}{\includegraphics[height=1.8cm]{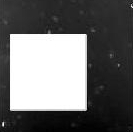}}
            & 
            \raisebox{-.5\height}{\includegraphics[height=1.8cm]{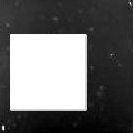}}            
            \\            
            \hline
            \textbf{FC-GANs Example Output:}
            & 
            \raisebox{-.5\height}{\includegraphics[height=1.8cm]{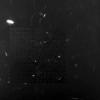}}
            & 
            \raisebox{-.5\height}{\includegraphics[height=1.8cm]{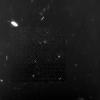}}            
            & 
            \raisebox{-.5\height}{\includegraphics[height=1.8cm]{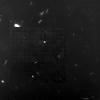}}
            & 
            \raisebox{-.5\height}{\includegraphics[height=1.8cm]{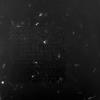}}
            &
            \raisebox{-.5\height}{\includegraphics[height=1.8cm]{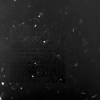}}
            & 
            \raisebox{-.5\height}{\includegraphics[height=1.8cm]{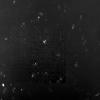}}
            & 
            \raisebox{-.5\height}{\includegraphics[height=1.8cm]{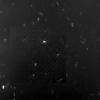}}            
            &
            \raisebox{-.5\height}{\includegraphics[height=1.8cm]{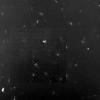}}
            & 
            \raisebox{-.5\height}{\includegraphics[height=1.8cm]{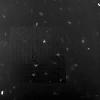}}
            & 
            \raisebox{-.5\height}{\includegraphics[height=1.8cm]{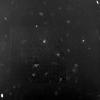}}            
            \\
            \hline
            \textbf{Saliency:}
            & 
            \raisebox{-.5\height}{\includegraphics[height=1.8cm]{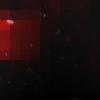}}
            & 
            \raisebox{-.5\height}{\includegraphics[height=1.8cm]{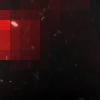}}            
            & 
            \raisebox{-.5\height}{\includegraphics[height=1.8cm]{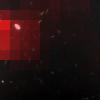}}
            & 
            \raisebox{-.5\height}{\includegraphics[height=1.8cm]{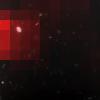}}
            &
            \raisebox{-.5\height}{\includegraphics[height=1.8cm]{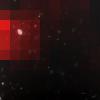}}
            & 
            \raisebox{-.5\height}{\includegraphics[height=1.8cm]{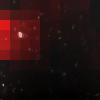}}
            & 
            \raisebox{-.5\height}{\includegraphics[height=1.8cm]{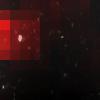}}            
            &
            \raisebox{-.5\height}{\includegraphics[height=1.8cm]{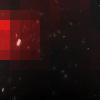}}
            & 
            \raisebox{-.5\height}{\includegraphics[height=1.8cm]{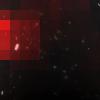}}
            & 
            \raisebox{-.5\height}{\includegraphics[height=1.8cm]{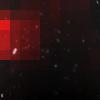}}            
            \\
             \hline
        \end{tabu}\egroup}
        \caption{Row 1 depicts a 10 frame sequence with an aggregate floating in the top left corner. FC-GANs are used to obscure the aggregate in frames 4 to 10 (Rows 2 and 3). By repeating this process and sliding the mask across the frames, we can compute the absolute differences in predictions, and are thereby able to compute a saliency matrix, which we subsequently apply to the ground truth frames (row 4).}
        \label{fig:FC-GANs:Completed_Region_vs_gt}
\end{table*}

\section{FC-GANs based Saliency Maps} \label{sec:verifcation} 

To verify that the 3D-ConvNets are detecting the anomalies found within the ER and CR evaluation sets we compute saliency maps using Frame-Completion GANs (FC-GANs). In this section we first discuss the implementation and training of the FC-GANs, before outlining our algorithm for computing the saliency maps. This is followed by a qualitative analysis of our saliency maps. In Section \ref{sec:Self-training} we incorporate the FC-GANs-based saliency maps into a multi-classifier voting system to automatically label additional training samples.  

\subsection{Frame-Completion GANs Training} 

As discussed in Section \ref{sec:background} we are using FC-GANs inspired by the image completion GANs from~\cite{iizuka2017globally} to compute our saliency maps. We train the FC-GANs using only \emph{AF} samples, meaning the filled in region is unlikely to contain anomalies. The generator receives the samples with masked frames as input. During training the location and dimensions of the mask are randomly selected. The inputs are subsequently processed by a fully convolutional network, trained to complete the masked region. 

\subsection{Computing Saliency Maps} 

We compute our saliency maps by applying a sliding mask to an input sample, using a trained FC-GANs generator to obtain completed frames. At each location we compute the absolute difference from the original prediction, allowing us to identify salient regions. Therefore, given a trained classifier $C$ and a generator $G$, we compute a saliency map as follows for a sample $X$. First $C$ will predict the probability $p$ that $X$ contains an anomaly. Subsequently we compute a saliency map $S$ by sliding a $h \times w$-pixel mask over the input frames, using $G$ to complete the blanked out region, feeding the completed frames to $C$, and observing the absolute difference between the probability $p'$ and $p$. The difference is added to corresponding saliency map cells that were masked within the input. Finally, a matrix $M$ is maintained to compute the number of times each cell within the saliency map is updated, which is used to obtain the average saliency score for each cell, as outlined in Algorithm \ref{alg:salmap}. 

\begin{algorithm}[h]
\caption{Computing a saliency map}
\label{alg:salmap}
  \begin{algorithmic}[1]
    \State \textbf{Input:} Classifier $C$, Generator $G$, Mask $h \times w$, Sample~$X$
    \State \textbf{Init:} Saliency map $S$, Counter matrix $M$, Stride $\eta$ 
    \State $p \gets C(X)$
    \For{$x=1$, $x=x+\eta$, while $x+w < width(X)$}
        \For{$y=1$, $y=y+\eta$, while $y+h < height(X)$} 
            \State $Y = G((X[x:x+w, y:y+h] = 0))$
            \State $S[x:x+w, y:y+h] \mathrel{+}= \abs{C(Y) - p}$
            \State $M[x:x+w, y:y+h] \mathrel{+}= 1$
        \EndFor        
    \EndFor
    \State Return $S \oslash M$
    \end{algorithmic}
\end{algorithm}

\subsection{Saliency Map Evaluation} 

As depicted in Table \ref{fig:FC-GANs:Completed_Region_vs_gt}, FC-GANs offer a means through which to remove anomalies and replacing the masked area with the type of suspensions one would expect in the evaluated product. Furthermore, adding the saliency map as a separate color channel to the original frames allows us to visualize the salient features within the input images and gain interesting insights, as depicted in Table \ref{tbl:saliency_maps}. First we observe that in frames with sufficient movement the classifier's predictions appear to be based on the anomalies, which can be distinguished from reflections, dust particles and the edges of the vials. Furthermore, via the saliency maps we can gain insights regarding the trajectory of anomalies that travel large distances, and they allow us to confirm that the classifiers are capable of distinguishing small aggregates from proteins. Through the saliency maps we can verify that 3D-ConvNets are able to identity anomalies of different shapes and sizes irrespective of location\footnote{We provide a link to a video of our saliency maps:  \url{https://youtu.be/S1IapmRl9H0}}. Saliency maps are therefore a first step towards providing a valuable tool to help visual inspectors interpret decisions made by the classifiers.

\begin{table*}
        \centering
        \resizebox{\textwidth}{!}{%
        \bgroup
        \footnotesize
        \tabulinesep=1.5mm
        \begin{tabu}{p{1.6cm}cccccccccc}
            \toprule
            \textbf{Time-step:} & \textbf{1} & \textbf{3} & \textbf{5} & \textbf{7} & \textbf{9} & \textbf{11} & \textbf{13} & \textbf{15} & \textbf{17} & \textbf{19} \\
            \toprule
            \multirow{2}{*}{\textbf{Example 1:}} 
            & 
            \raisebox{-.5\height}{\includegraphics[height=1.8cm]{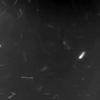}}
            & 
            \raisebox{-.5\height}{\includegraphics[height=1.8cm]{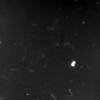}}
            & 
            \raisebox{-.5\height}{\includegraphics[height=1.8cm]{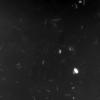}}
            & 
            \raisebox{-.5\height}{\includegraphics[height=1.8cm]{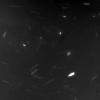}}
            &
            \raisebox{-.5\height}{\includegraphics[height=1.8cm]{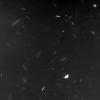}}
            & 
            \raisebox{-.5\height}{\includegraphics[height=1.8cm]{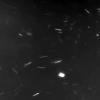}}
            & 
            \raisebox{-.5\height}{\includegraphics[height=1.8cm]{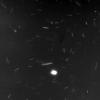}}
            &
            \raisebox{-.5\height}{\includegraphics[height=1.8cm]{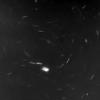}}
            & 
            \raisebox{-.5\height}{\includegraphics[height=1.8cm]{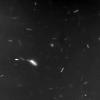}}
            & 
            \raisebox{-.5\height}{\includegraphics[height=1.8cm]{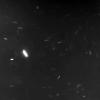}}            
            \\
            &
            \raisebox{-.5\height}{\includegraphics[height=1.8cm]{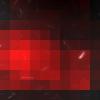}}
            & 
            \raisebox{-.5\height}{\includegraphics[height=1.8cm]{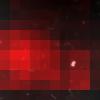}}
            & 
            \raisebox{-.5\height}{\includegraphics[height=1.8cm]{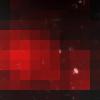}}
            & 
            \raisebox{-.5\height}{\includegraphics[height=1.8cm]{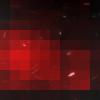}}
            &
            \raisebox{-.5\height}{\includegraphics[height=1.8cm]{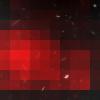}}
            & 
            \raisebox{-.5\height}{\includegraphics[height=1.8cm]{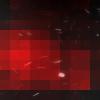}}
            & 
            \raisebox{-.5\height}{\includegraphics[height=1.8cm]{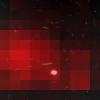}}
            &
            \raisebox{-.5\height}{\includegraphics[height=1.8cm]{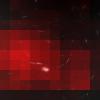}}
            & 
            \raisebox{-.5\height}{\includegraphics[height=1.8cm]{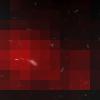}}
            & 
            \raisebox{-.5\height}{\includegraphics[height=1.8cm]{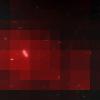}}            
            \\
            \hline              
            \multirow{2}{*}{\textbf{Example 2:}} 
            & 
            \raisebox{-.5\height}{\includegraphics[height=1.8cm]{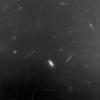}}
            & 
            \raisebox{-.5\height}{\includegraphics[height=1.8cm]{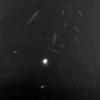}}
            & 
            \raisebox{-.5\height}{\includegraphics[height=1.8cm]{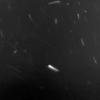}}
            & 
            \raisebox{-.5\height}{\includegraphics[height=1.8cm]{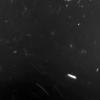}}
            &
            \raisebox{-.5\height}{\includegraphics[height=1.8cm]{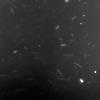}}
            & 
            \raisebox{-.5\height}{\includegraphics[height=1.8cm]{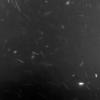}}
            & 
            \raisebox{-.5\height}{\includegraphics[height=1.8cm]{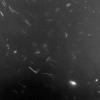}}
            &
            \raisebox{-.5\height}{\includegraphics[height=1.8cm]{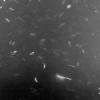}}
            & 
            \raisebox{-.5\height}{\includegraphics[height=1.8cm]{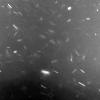}}
            & 
            \raisebox{-.5\height}{\includegraphics[height=1.8cm]{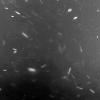}}            
            \\
            & 
            \raisebox{-.5\height}{\includegraphics[height=1.8cm]{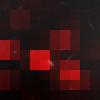}}
            & 
            \raisebox{-.5\height}{\includegraphics[height=1.8cm]{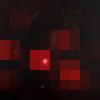}}
            & 
            \raisebox{-.5\height}{\includegraphics[height=1.8cm]{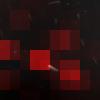}}
            & 
            \raisebox{-.5\height}{\includegraphics[height=1.8cm]{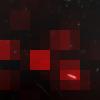}}
            &
            \raisebox{-.5\height}{\includegraphics[height=1.8cm]{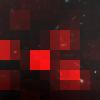}}
            & 
            \raisebox{-.5\height}{\includegraphics[height=1.8cm]{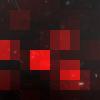}}
            & 
            \raisebox{-.5\height}{\includegraphics[height=1.8cm]{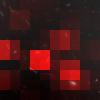}}
            &
            \raisebox{-.5\height}{\includegraphics[height=1.8cm]{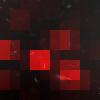}}
            & 
            \raisebox{-.5\height}{\includegraphics[height=1.8cm]{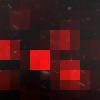}}
            & 
            \raisebox{-.5\height}{\includegraphics[height=1.8cm]{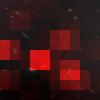}}            
            \\
            \hline              
            \multirow{2}{*}{\textbf{Example 3:}} 
            & 
            \raisebox{-.5\height}{\includegraphics[height=1.8cm]{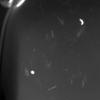}}
            & 
            \raisebox{-.5\height}{\includegraphics[height=1.8cm]{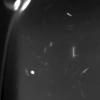}}
            & 
            \raisebox{-.5\height}{\includegraphics[height=1.8cm]{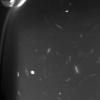}}
            & 
            \raisebox{-.5\height}{\includegraphics[height=1.8cm]{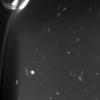}}
            &
            \raisebox{-.5\height}{\includegraphics[height=1.8cm]{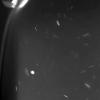}}
            & 
            \raisebox{-.5\height}{\includegraphics[height=1.8cm]{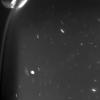}}
            & 
            \raisebox{-.5\height}{\includegraphics[height=1.8cm]{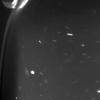}}
            &
            \raisebox{-.5\height}{\includegraphics[height=1.8cm]{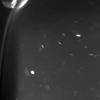}}
            & 
            \raisebox{-.5\height}{\includegraphics[height=1.8cm]{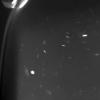}}
            & 
            \raisebox{-.5\height}{\includegraphics[height=1.8cm]{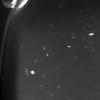}}            
            \\
            & 
            \raisebox{-.5\height}{\includegraphics[height=1.8cm]{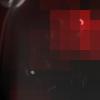}}
            & 
            \raisebox{-.5\height}{\includegraphics[height=1.8cm]{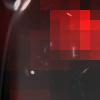}}
            & 
            \raisebox{-.5\height}{\includegraphics[height=1.8cm]{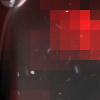}}
            & 
            \raisebox{-.5\height}{\includegraphics[height=1.8cm]{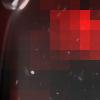}}
            &
            \raisebox{-.5\height}{\includegraphics[height=1.8cm]{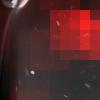}}
            & 
            \raisebox{-.5\height}{\includegraphics[height=1.8cm]{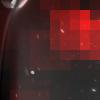}}
            & 
            \raisebox{-.5\height}{\includegraphics[height=1.8cm]{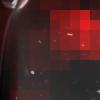}}
            &
            \raisebox{-.5\height}{\includegraphics[height=1.8cm]{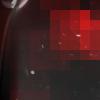}}
            & 
            \raisebox{-.5\height}{\includegraphics[height=1.8cm]{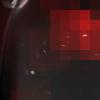}}
            & 
            \raisebox{-.5\height}{\includegraphics[height=1.8cm]{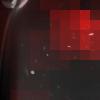}}            
            \\
            \hline
            \multirow{2}{*}{\textbf{Example 4:}} 
            & 
            \raisebox{-.5\height}{\includegraphics[height=1.8cm]{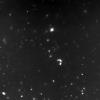}}
            & 
            \raisebox{-.5\height}{\includegraphics[height=1.8cm]{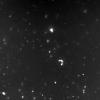}}
            & 
            \raisebox{-.5\height}{\includegraphics[height=1.8cm]{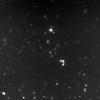}}
            & 
            \raisebox{-.5\height}{\includegraphics[height=1.8cm]{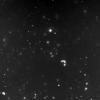}}
            &
            \raisebox{-.5\height}{\includegraphics[height=1.8cm]{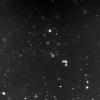}}
            & 
            \raisebox{-.5\height}{\includegraphics[height=1.8cm]{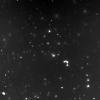}}
            & 
            \raisebox{-.5\height}{\includegraphics[height=1.8cm]{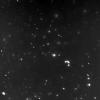}}
            &
            \raisebox{-.5\height}{\includegraphics[height=1.8cm]{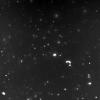}}
            & 
            \raisebox{-.5\height}{\includegraphics[height=1.8cm]{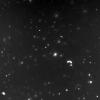}}
            & 
            \raisebox{-.5\height}{\includegraphics[height=1.8cm]{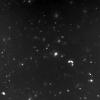}}            
            \\            
            & 
            \raisebox{-.5\height}{\includegraphics[height=1.8cm]{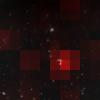}}
            & 
            \raisebox{-.5\height}{\includegraphics[height=1.8cm]{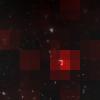}}
            & 
            \raisebox{-.5\height}{\includegraphics[height=1.8cm]{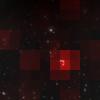}}
            & 
            \raisebox{-.5\height}{\includegraphics[height=1.8cm]{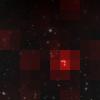}}
            &
            \raisebox{-.5\height}{\includegraphics[height=1.8cm]{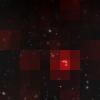}}
            & 
            \raisebox{-.5\height}{\includegraphics[height=1.8cm]{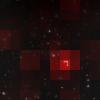}}
            & 
            \raisebox{-.5\height}{\includegraphics[height=1.8cm]{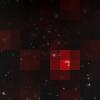}}
            &
            \raisebox{-.5\height}{\includegraphics[height=1.8cm]{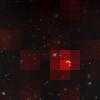}}
            & 
            \raisebox{-.5\height}{\includegraphics[height=1.8cm]{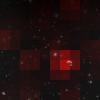}}
            & 
            \raisebox{-.5\height}{\includegraphics[height=1.8cm]{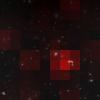}}            
            \\
            \hline
        \end{tabu}\egroup}
        \caption{For each example the red areas illustrate the saliency (bottom row) for the inputs (top row). Examples 1 \& 2 depict the trajectory of two aggregates. Example 3 shows the maps can be used to verify that classifications are not based on confounding factors such as reflections (bottom left) and edges (upper left). In Example 4 we see the network can distinguish a small aggregate from a good particle.} 
        \label{tbl:saliency_maps}
    \end{table*}

\begin{table*}
    \centering
    \resizebox{\textwidth}{!}{%
    \bgroup
    \footnotesize
    \tabulinesep=1.5mm
    \begin{tabu}{|p{1.2cm}||c|c|c|c|}
        \hline
        \textbf{Eval. Set:} & \textbf{All} & \textbf{Anomaly Free (AF)} & \textbf{Easy Rejects (ER)} & \textbf{Challenging Rejects} \\
        \hline
        \hline
        \textbf{Accuracy:}
        & 
        \raisebox{-.5\height}{\includegraphics[width=0.24\textwidth]{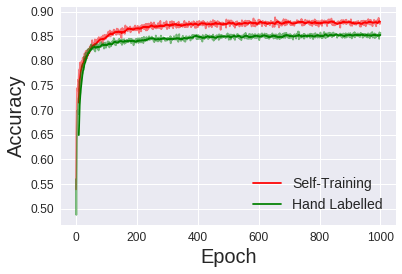} }
        & 
        \raisebox{-.5\height}{\includegraphics[width=0.24\textwidth]{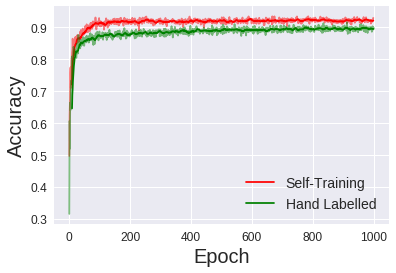} }
        & 
        \raisebox{-.5\height}{\includegraphics[width=0.24\textwidth]{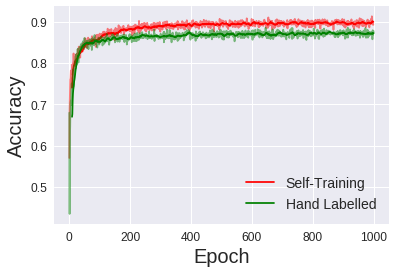} }
        & 
        \raisebox{-.5\height}{\includegraphics[width=0.24\textwidth]{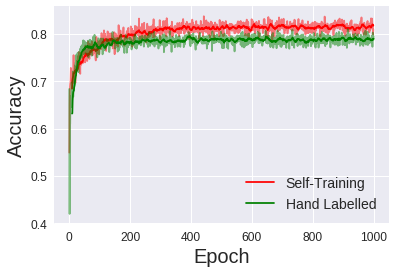} }
        \\
        \hline
        \textbf{Loss:}
        & 
        \raisebox{-.5\height}{\includegraphics[width=0.24\textwidth]{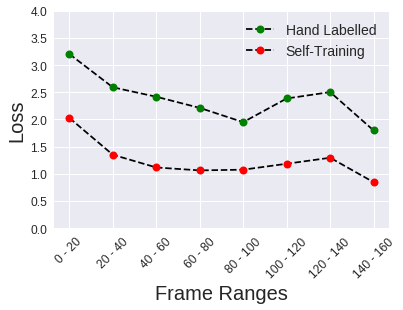} }
        & 
        \raisebox{-.5\height}{\includegraphics[width=0.24\textwidth]{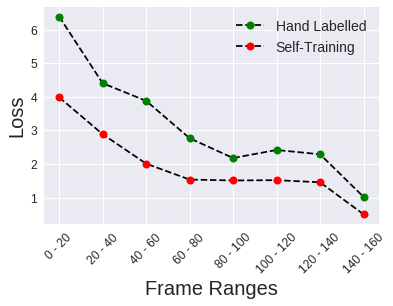} }
        & 
        \raisebox{-.5\height}{\includegraphics[width=0.24\textwidth]{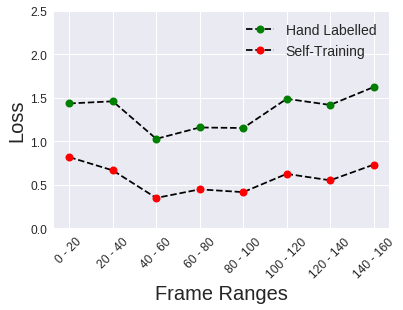} }
        & 
        \raisebox{-.5\height}{\includegraphics[width=0.24\textwidth]{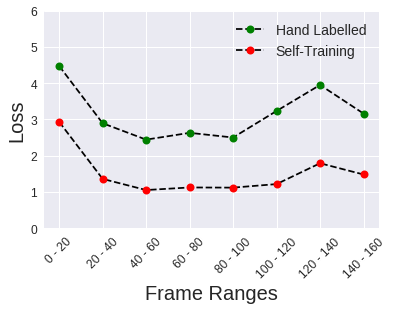} }
        \\
        \hline            
    \end{tabu}\egroup}
    \caption{In the first row we depict the average accuracy achieved for each evaluation set. We observe that classifiers optimized with the self-training dataset outperforms those trained with the smaller hand-labelled set. In the second row we provide a breakdown of the average loss according to the frame range during which the evaluation sample was extracted.} 
    \label{tbl:accuracy}
\end{table*}

\section{Self-training} \label{sec:Self-training}

\begin{figure}[H]
\centering
\includegraphics[width=0.81\columnwidth]{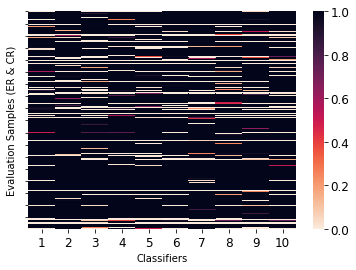} 
\caption{The heatmap above illustrates that the classifiers often disagree regarding a vial's status, with darker areas indicating that samples received a higher positive prediction.}
\label{fig:classifier_disagreement}      
\end{figure}

Despite being trained with the same dataset, upon plotting the predictions of each classifier for ER and CR evaluation samples in a heatmap (Figure \ref{fig:classifier_disagreement}), we observe that the classifiers often disagree. Therefore initializing each network using a unique seed value and stochastic sampling are a sufficient condition for convergence upon different optima. We observe that networks having different strengths can enable the construction of a diverse dataset during automatic labelling. We assign \emph{positive} (containing anomalies) and \emph{negative} (anomaly-free) labels using \emph{strict} and \emph{lenient} voting conditions, respectively, with the following intuition:

\begin{enumerate}
    \item We observe a sample is likely to be anomaly free when classified as \emph{negative} under \emph{strict} voting conditions, where a positive prediction is triggered when a small subset of classifiers believe there is an anomaly. 
    \item For positive vials we observe that false-positives can be minimized under \emph{lenient} voting conditions, where a subset of classifiers must agree both on the likelihood of a sample containing an anomaly and the location.
\end{enumerate}

We use the FC-GANs based saliency maps to measure the agreement between classifiers under lenient voting conditions. A \emph{positive} label is assigned to samples only when $n > 1$ classifiers predict with above $0.8$ certainty that a vial is positive, and with a median pairwise $L_2$ distance between saliency maps that is less than $20.0$. We chose these values by applying the lenient voting condition to a set of samples derived from anomaly free vials that were previously not included in the training or evaluation sets. This allows us to keep the number of false-positives to $4.5\%$, while assigning \emph{positive} labels to $108,944$ ($35.45\%$ of) unlabelled samples. We subsequently use the strict labelling condition to obtain an additional $108,944$ \emph{negative} samples. Prior to automatically assigning labels we exclude samples from the less indicative frame ranges, only keeping sequences starting at $40 \leq t \leq 100$. We subsequently augment our dataset, adding $217,888$ self-training samples to our 14,000 hand-labelled samples, and train an additional ten randomly initialized 3D-ConvNets. 

\begin{figure}[h]
\centering
\subfigure[Positive Samples]{\label{fig:roc:rej}\includegraphics[width=0.9\columnwidth]{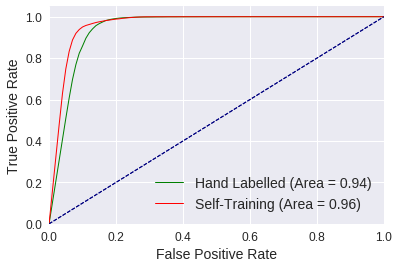}}
\subfigure[Negative Samples]{\label{fig:roc:acc}\includegraphics[width=0.9\columnwidth]{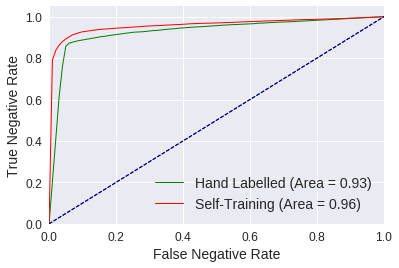}}
\caption{AUROC Plots}
\label{fig:ROC:3D-CNN}
\end{figure}

The ConvNets optimized with the self-training dataset significantly outperform those using only hand-labelled samples, as evident from the time-series plots depicted in the first row of Table \ref{tbl:accuracy}. Furthermore, the frame range plots illustrating the mean cross entropy loss in the second row show a decrease in error across all evaluation sets. This translates to an increase in prediction accuracy for challenging rejects, achieving 85.1\% between frame range 40 to 100. Furthermore, the overall accuracy for evaluation sets of 88.7\% between frames 80 and 100 increases to 90.5\%. Finally, we compute the AUROC for both approaches using samples between frame ranges 40 and 120. Self-training achieves an AUROC of 0.96 for positive and negative samples, compared to 0.94 and 0.93 when only using the hand-labelled dataset (see Figure \ref{fig:ROC:3D-CNN}). This evidence supports that disagreeing classifiers can be used under \emph{strict} and \emph{lenient} classification conditions, to automatically label and add samples to a dataset while providing sufficient sample variance for learning improved models.
\section{Future Work} \label{sec:FutureWork}

We have successfully demonstrated the potential of deep learning for the automated inspection of opaque liquid vaccines and are currently working with HAL and other partners to develop this into a commercial solution. As evident from our self-training experiment in Section~\ref{sec:Self-training}, the 3D-ConvNets can benefit from using a richer dataset for optimization, ideally using samples derived from a diverse set of vials via a multi-camera conveyor belt system. Other interesting avenues for future research include: 

\begin{itemize}
    \item Robotics research suggests that additional sensory input obtained from being able to manipulate an object via interactive perception can lead to improved classification~\cite{bohg2017interactive}. We are therefore investigating the feasibility of handing over manual control of the vials to an agent that can determine the inspection pose, regions of interest and vial angle, while also being able to re-sample a vial if sampling leads to an uncertain prediction. We note that human visual inspectors often use the option of re-examining a vial. 

    \item In the interest of reducing training times and gathering multiple-runs we down-sampled our frames during pre-processing. However, arguably some useful details are lost during this step, and it would be reasonable to expect a further improvement in accuracy if the classifiers were to be trained using larger frames. An increase in detail should help with the detection of smaller boundary samples, where even human inspectors reach their limit. 

    \item For the experiments outlined in Section \ref{sec:Self-training} we trained a new set of randomly initialized classifiers using our self-training dataset. However, we observe that re-training a set of pre-trained networks using a larger dataset could reduce the amount of time required to achieve convergence. Indeed, limiting the amount of training time that deep learning architectures require and reducing delays to the production work flow is critical within an industrial setting~\cite{schnieders2019fully,schnieders2018fast}. Therefore, evaluating to what extent optimizing pre-trained networks can enable a faster convergence in this context, without having to compromise on accuracy, represents an important avenue for future work in this area.   

    \item Since this technology will be a component of a critical system we shall look further into verification and interpretability, building on our work for computing saliency maps to help interpret the decisions made by classifiers. 
    
    \item While this paper focuses on suspensions, we are currently looking to obtain datasets to evaluate the general applicability of the techniques discussed towards other formats, e.g., clear solutions. 
    
\end{itemize}

\section{Conclusion} \label{sec:conclusion}

We have provided evidence that deep learning can be used to automate the process of visually inspecting \emph{opaque} liquid pharmaceutical vaccines containing suspensions. While our work shows the benefits of training classifiers using an augmented dataset obtained via a novel self-training approach, we also provide a comparison against human judgement, namely the ground truth labelling. Here it is worth noting that despite deep learning being widely used in the fields of computer vision and biological image processing, the trained networks rarely match human performance (which in our case would mean 100\% accuracy). However, while the performance and availability of human inspectors may vary, e.g., due to tiredness, sickness, vacations, etc, automated systems can operate indefinitely while delivering consistent, competitive performance that almost matches humans. 

To summarize our contributions: \\
\noindent{\textbf{1)}} We outline a process for recording a video dataset of liquid vaccine samples containing suspension. We use a hand built automated vial rotator (AVR) to standardise the recording of liquid vaccines supplied by \anonymize{the HAL Allergy Group} and obtain recordings for our dataset. To improve the quality of our dataset we manually labelled 14,000 training and 6,000 evaluation samples, with each sample consisting of $20$ frames of $100 \times 100$ pixels.\\ 
\noindent{\textbf{2)}} Using this dataset we train ten randomly initialized 3D-ConvNets, where upon computing the AUROC we observe scores of 0.94 and 0.93 for positive (anomaly containing) and negative (anomaly-free) samples, respectively.\\   
\noindent{\textbf{3)}} We introduce an algorithm which uses Frame-Completion GANs to identify salient regions within inputs, and subsequently use this method to verify that the classifiers are learning to identify anomalies within the vials. \\
\noindent{\textbf{4)}} Given the small size of our dataset, we use self-training, automatically labelled $217,888$ 20-frame samples. To reduce the likelihood of noisy labels we use a voting system that also makes use of the FC-GANs based saliency maps to determine when classifiers are in agreement regarding an anomaly's location. Classifiers trained with the augmented dataset achieve AUROC scores of 0.96 for both positive and negative samples, improving on the benchmarks set by 3D-ConvNets using our unaugmented dataset (See Table \ref{tbl:accuracy} and Figure~\ref{fig:ROC:3D-CNN}).    

\ack This paper is dedicated to the memory of our wonderful colleague and friend Benjamin Schnieders, a bright young scientist, who recently passed away. We thank the HAL Allergy Group for partially funding the PhD of Gregory Palmer and gratefully acknowledge the support of NVIDIA Corporation with the donation of the Titan X Pascal GPU that enabled this research.

\balance
\bibliography{ecai}
\end{document}